\documentclass[conference]{IEEEtran}
\IEEEoverridecommandlockouts
\usepackage{cite}
\usepackage{amsmath,amssymb,amsfonts}
\usepackage{booktabs} 
\usepackage{algorithmic}
\usepackage{graphicx}
\usepackage{textcomp}
\usepackage{xcolor}
\usepackage{algorithm}
\usepackage{algorithmic}
\usepackage{amsmath} 
\usepackage{amssymb} 
\usepackage{booktabs}
\usepackage{pifont} 
\usepackage{adjustbox}
\usepackage{mathtools}

\usepackage{caption}
\usepackage{multirow}
\usepackage{url}
\usepackage{authblk}

\usepackage{algorithm}

\newcommand{\xmark}{\ding{55}} 
\def\BibTeX{{\rm B\kern-.05em{\sc i\kern-.025em b}\kern-.08em
    T\kern-.1667em\lower.7ex\hbox{E}\kern-.125emX}}

\begin{document}

\title{IMAN: An Adaptive Network for Robust NPC Mortality Prediction with Missing Modalities}

\author[1]{Yejing Huo}
\author[1*]{Guoheng Huang}
\author[1]{Lianglun Cheng}
\author[1]{Jianbin He}
\author[25]{Xuhang Chen}
\author[3]{Xiaochen Yuan}
\author[4]{\authorcr Guo Zhong}
\author[5]{Chi-Man Pun}
\affil[1]{School of Computer Science and Technology, Guangdong University of Technology}
\affil[2]{School of Computer Science and Engineering, Huizhou University}
\affil[3]{Faculty of Applied Sciences, Macao Polytechnic University}
\affil[4]{School of Information Science and Technology, Guangdong University of Foreign Studies}
\affil[5]{Department of Computer and Information Science, University of Macau}


\maketitle

\begin{abstract}
Accurate prediction of mortality in nasopharyngeal carcinoma (NPC), a complex malignancy particularly challenging in advanced stages, is crucial for optimizing treatment strategies and improving patient outcomes. However, this predictive process is often compromised by the high-dimensional and heterogeneous nature of NPC-related data, coupled with the pervasive issue of incomplete multi-modal data, manifesting as missing radiological images or incomplete diagnostic reports. Traditional machine learning approaches suffer significant performance degradation when faced with such incomplete data, as they fail to effectively handle the high-dimensionality and intricate correlations across modalities. Even advanced multi-modal learning techniques like Transformers struggle to maintain robust performance in the presence of missing modalities, as they lack specialized mechanisms to adaptively integrate and align the diverse data types, while also capturing nuanced patterns and contextual relationships within the complex NPC data. To address these problem, we introduce IMAN: an adaptive network for robust NPC mortality prediction with missing modalities. IMAN features three integrated modules: the Dynamic Cross-Modal Calibration (DCMC) module employs adaptive, learnable parameters to scale and align medical images and field data; the Spatial-Contextual Attention Integration (SCAI) module enhances traditional Transformers by incorporating positional information within the self-attention mechanism, improving multi-modal feature integration; and the Context-Aware Feature Acquisition (CAFA) module adjusts convolution kernel positions through learnable offsets, allowing for adaptive feature capture across various scales and orientations in medical image modalities. Extensive experiments on our proprietary NPC dataset demonstrate IMAN's robustness and high predictive accuracy, even with missing data. Compared to existing methods, IMAN consistently outperforms in scenarios with incomplete data, representing a significant advancement in mortality prediction for medical diagnostics and treatment planning. Our code is available at \url{https://github.com/king-huoye/BIBM-2024/tree/master}.
\end{abstract}


\begin{figure}[ht]
    \centering
    \includegraphics[width=\columnwidth]{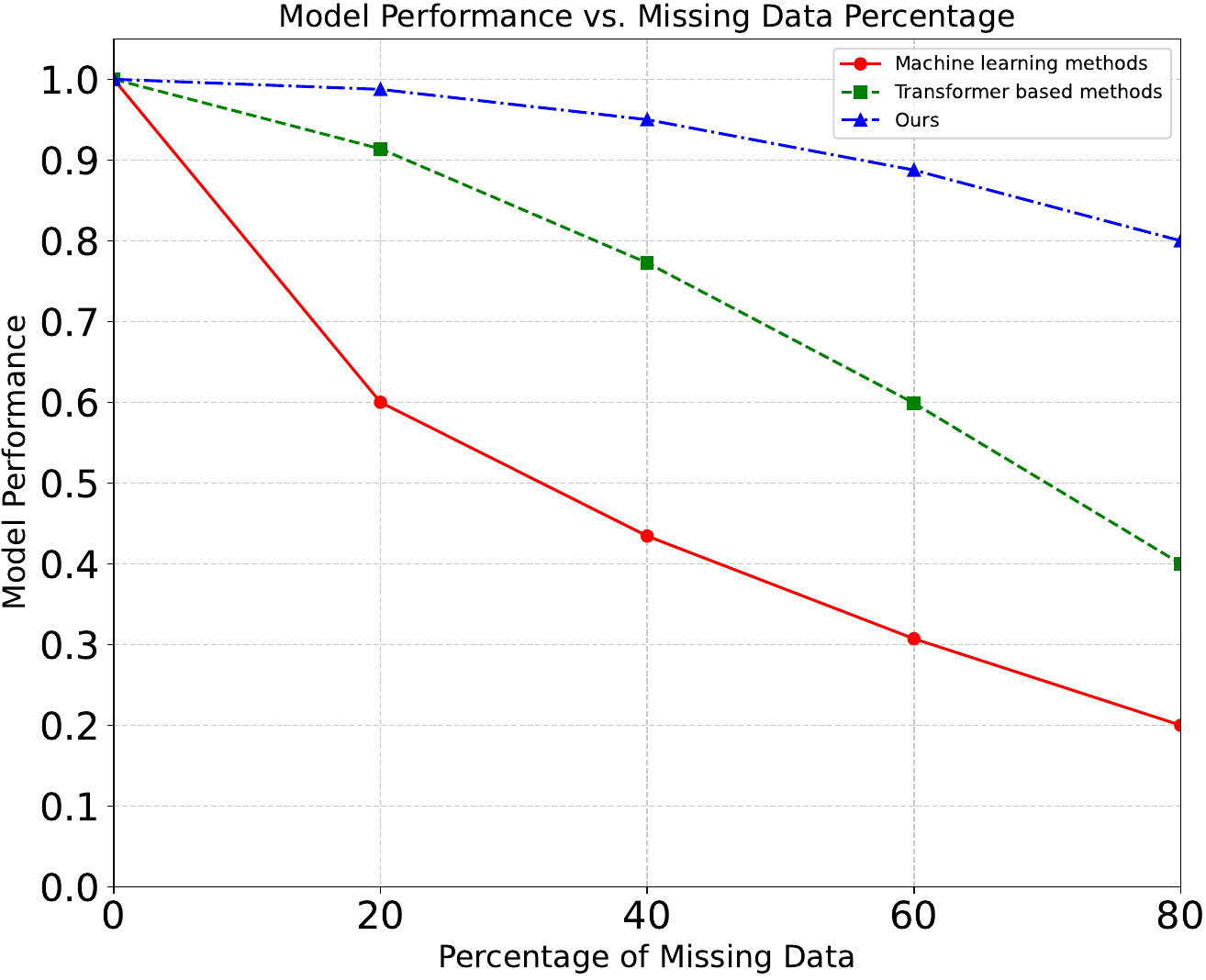}
    \caption{Comparison of performance at different levels of missingness. 
    }
    
    \label{fig:motivation}
\end{figure}

\section{Introduction}

Medical diagnostics and prognostics for complex diseases like nasopharyngeal carcinoma (NPC) heavily rely on comprehensive data analysis. 
NPC early detection and accurate prognosis are vital for timely, effective treatment, significantly improving clinical outcomes and quality of life \cite{bib4,py1,py2,py4,py5,py6}. Consequently, there is an urgent need for accurate predictive models that can effectively integrate diverse data modalities to enhance early diagnosis, reduce NPC mortality, and guide optimal treatment plans.

To address this pressing need, early traditional machine learning techniques like Lasso \cite{bib6} have been employed, significantly improving prognostic accuracy for advanced-stage NPC patients. However, these methods often struggle due to their inability to effectively handle data heterogeneity,  and missing modalities, severely impacting their performance (as shown in Figure \ref{fig:motivation}). The rise of deep learning has facilitated the development of multi-modal approaches that leverage powerful neural networks to integrate diverse modalities \cite{chen1,py3}. Early deep learning methods utilized matrix completion \cite{bib7} to address missing modalities, but suffered from high computational complexity and performance degradation as the number of modalities increased. More recent studies have explored tensor-based modal distillation \cite{bib8} and Transformer-based feature reconstruction \cite{bib9, bib10} to tackle these challenges. Nevertheless, existing methods commonly assume a single missing modality and falter when confronted with highly heterogeneous and incomplete data contexts involving multiple missing modalities, as exemplified by Lee et al.'s \cite{bib10} multi-modal Transformer with prompt learning \cite{chen3,chen5}. These limitations highlight the critical need for a more robust solution capable of effectively handling incomplete multi-modal data in nasopharyngeal carcinoma mortality prediction tasks.

To address the above shortcomings, we propose IMAN, an adaptive network for reliable NPC death prediction under modality missing. IMAN features three key modules: Dynamic Cross-Modal Calibration (DCMC), Spatial-Contextual Attention Integration (SCAI), and Context-Aware Feature Acquisition (CAFA). The DCMC module uses learnable parameters to adaptively scale and align medical images and text data, enhancing the normalization of heterogeneous inputs. The SCAI module improves feature fusion by incorporating positional information into the self-attention mechanism, enabling the detection of subtle patterns in medical images. The CAFA module adjusts convolution kernel positions through learned offsets, allowing adaptive feature capture across various scales and orientations. Extensive experiments on our NPC dataset demonstrate IMAN's robustness and high predictive accuracy, even with missing data. This integrated approach ensures more consistent and accurate treatment outcome predictions, representing a significant advancement in NPC diagnosis and treatment planning.

The contributions of this work are summarized as follows:
\begin{itemize}
    \item We propose IMAN, an Incomplete Modality Adaptive Network, to address the poor performance of models with incomplete modalities in predicting nasopharyngeal carcinoma (NPC) outcomes. IMAN consists of three core modules: DCMC, SCAI, and CAFA.
    \item The DCMC module employs learnable parameters to adaptively scale and align medical images and field data, enhancing the normalization and alignment of heterogeneous inputs. The SCAI module integrates positional information into the self-attention mechanism of Transformers, improving the fusion of multi-modal features and enabling the model to detect subtle patterns in medical data. The CAFA module adjusts convolution kernel positions through learned offsets, allowing adaptive feature capture across different scales and orientations in various medical image modalities.
    \item Extensive experiments demonstrate that IMAN performs exceptionally well on the proprietary NPC datasets.
\end{itemize}

\section{Method}

\begin{figure}[ht]
    \centering
    \includegraphics[width=\columnwidth]{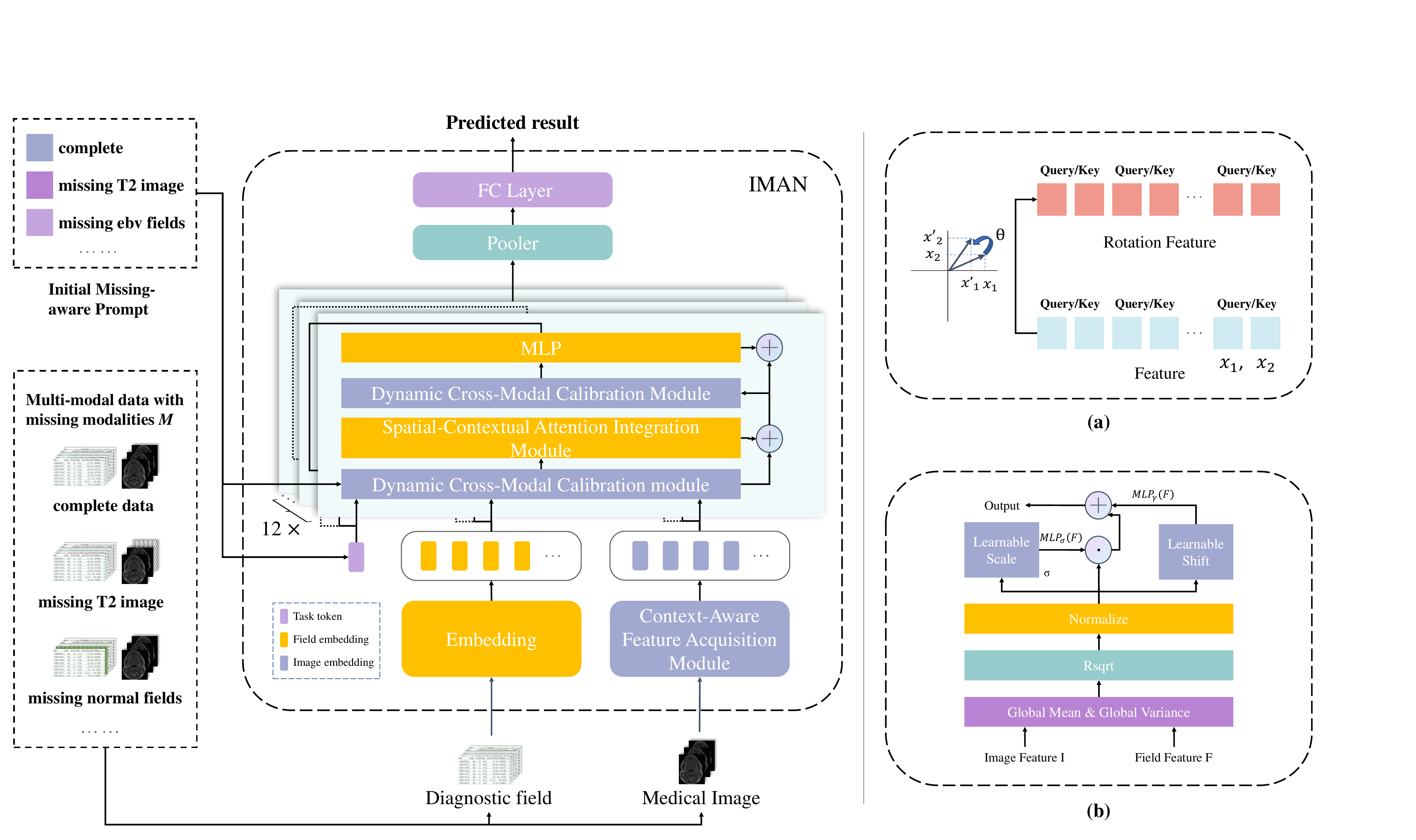}
    \caption{The overview of IMAN. 
    }
    \label{fig:flow}
\end{figure}

\subsection{The overview of our method}

\begin{algorithm}
\caption{Algorithmic flow of Context-Aware Feature Acquisition module.}
\label{algorithm:AFE}
\begin{algorithmic}[1]
\STATE \textbf{Input:} \texttt{Kernel size:} \texttt{Num\_param}, \texttt{Data type:} \texttt{Dtype}
\STATE \textbf{Output:} \texttt{Initial sampling coordinates} \texttt{$P_n$}

\STATE \textbf{Step 1: Compute base integer and row number.}
\STATE \texttt{Base\_int} $\gets$ \texttt{round(sqrt(Num\_param))}
\STATE \texttt{Row\_number} $\gets$ \texttt{Num\_param} \texttt{//} \texttt{Base\_int}
\STATE \texttt{Mod\_number} $\gets$ \texttt{Num\_param} \% \texttt{Base\_int}

\STATE \textbf{Step 2: Obtain and flatten regular kernel coordinates.}\STATE $P_{n_x}, P_{n_y} \gets \text{meshgrid}(\texttt{Row\_number}, \texttt{Base\_int})$
\STATE \texttt{Flatten} \texttt{$P_{n_{x}}$} and \texttt{$P_{n_{y}}$}

\IF{\texttt{Mod\_number} $>$ 0}
    \STATE \textbf{Step 3: Include additional coordinates for irregular kernels.}
    \STATE \texttt{Extend \texttt{Row\_number} by 1, include extra \texttt{Mod\_number} positions}
    \STATE \texttt{Flatten and concatenate with existing coordinates}
\ENDIF

\STATE \textbf{Step 4: Combine and reshape coordinates.}
\STATE \texttt{$P_n$} $\gets$ concatenate(\texttt{$P_{n_{x}}$}, \texttt{$P_{n_{y}}$})
\STATE \texttt{Reshape and cast to} \texttt{Dtype}

\RETURN \texttt{$P_n$}
\end{algorithmic}
\end{algorithm}

The IMAN network follows a structured design, beginning with pre-training a missing-aware prompt module using diverse data to enhance predictive power across various scenarios. This pre-trained module generates task-specific tokens based on the presence or absence of different modalities, guiding the attention of the Transformer towards relevant information. Based on the ViLT architecture \cite{bib24}, IMAN transforms data fields into embedded tokens and uses the CAFA module to adjust convolutional kernels, capturing features across scales and orientations for multi-modal information. The pre-trained module generates task tokens with cues based on inputs, which are added to the sequence of field and image tokens, guiding the Transformer's attention mechanism to address missing modality issues and reduce fine-tuning needs. The combined inputs pass through 12 Transformer layers, with the DCMC module dynamically calibrating data sizes and the SCAI module incorporating positional information into the self-attention mechanism for enhanced feature fusion and performance. The Transformer's outputs are then processed through pooling and fully-connected layers for classification. Detailed functions and roles of each component are discussed in subsequent sections.

\subsection{Dynamic Cross-Modal Calibration Module}

Inspired by the migration style task \cite{Tikhonova2023TextSA} that transforms image styles to content images in feature space through feature statistics, we propose the DCMC module. DCMC introduces dynamic and learnable parameterization for normalization. It employs sets of learnable scaling parameters $\sigma$ and shifting parameters $\gamma$, optimized during training to match the statistical properties of input data. These parameters adaptively scale the feature maps of medical images and field data, promoting a consistent representation across modalities. The mathematical formulations are as follows:
\begin{equation}
    \mathrm{DCMC}(I,\mathbf{F})=\sigma(\mathbf{F})\odot\left(\frac{I-\mu(I)}{\sigma(I)}\right)+\gamma(\mathbf{F})
\end{equation}
\begin{equation}
\sigma(\mathbf{F})=\mathrm{MLP}_\sigma(\mathbf{F}),\quad\gamma(\mathbf{F})=\mathrm{MLP}_\gamma(\mathbf{F})
\end{equation}
where $\sigma$ and $\gamma$ are linearly transformed by two MLPs, respectively, from the field features F. The DCMC takes as input the image features I and the field features F and adapts the mean and standard deviation of I according to F, allowing I to adaptively match different F. In this way, the shape information contained in the image features can be injected into the fields while preserving their spatial structure.

IMAN's adaptive mechanism enhances its ability to synchronize diverse medical data, boosting operational efficiency in complex, high-dimensional datasets like nasopharyngeal carcinoma studies. The network's capability to handle varied data types strengthens its diagnostic and prognostic performance in medical imaging and treatment planning. The DCMC module significantly improves IMAN's capacity to process multi-modal data, crucial in scenarios with incomplete medical images, ensuring accurate diagnostic predictions without compromising performance.

\subsection{Spatial-Contextual Attetion Integration Module}

Traditional Transformer models face challenges with high-dimensional multi-modal data in contexts like predicting mortality in nasopharyngeal carcinoma (NPC), especially when modalities are missing. Addressing these issues, we introduce the SCAI module, inspired by Su et. al  \cite{Su_Lu_Pan_Wen_Liu_2021}. This module incorporates rotation-based methods within the self-attention mechanism, enhancing the modeling of relative positions crucial for spatial relationships in medical images and sequential patterns in diagnostic reports \cite{chen2,chen4}. Unlike fixed position embeddings, rotational encoding provides flexibility to manage varying sequence lengths and absent modalities. By preserving semantic content and improving the model's capability to detect subtle patterns across modalities, the SCAI module significantly enhances feature fusion effectiveness in complex, high-dimensional NPC data, facilitating better diagnosis and treatment planning.

Initially, we establish a sequence of features with a length of N (as the operations for field and image features are analogous in the process of SCAI, the symbol P in the subsequent equation denotes various operations for different eigenvectors under different circumstances): \(F_N=\{W_i\}_{i=1}^{N}\). Within this sequence,  \(w_i\)represents the i-th token of the input sequence. The embedding corresponding to the input sequence  \(F_N\) is denoted as  \(E_N=\{x_i\}_{i=1}^{N}\), where \(x_i\) signifies the d-dimensional embedding vector associated with the i-th token \(w_i\).

Prior to engaging in self-attention operations, we employ the feature embedding vectors to derive the query (q), key (k), and value (v) vectors, while also integrating the pertinent positional information. 

In this context, \(q_s\) denotes the query vector for the s-th token, which has incorporated the positional information s into its corresponding feature vector \(x_s\). Similarly, \(k_t\) and \(v_t\) represent the key and value vectors, respectively, for the t-th token, with the positional information t seamlessly integrated into the feature vector \(x_t\).

Subsequently, the objective is to compute the self-attention output for the s-th feature embedding vector \(x_s\). This process entails determining the attention scores between the query vector \(q_s\) and all other key vectors \(k_t\). Each attention score is then multiplied by its corresponding value vector \(v_t\). The final step involves aggregating these products through summation to derive the resultant output vector \(o_s\).

Moving forward, to capitalize on the relative positional relationships among the tokens in question, we introduce a function, denoted as $g$, which encapsulates the dot product operation between the query vector \(q_s\)  and the key vector \(k_t\). The function $g$ is designed to take into account the word embedding vectors \(x_s\) and \(x_t\), along with the relative positional difference s-t, thereby enabling the model to discern the contextual significance of each token's position within the sequence.
\begin{equation}
    <P_q(x_s,s),P_k(x_t,t)>=g(x_s,x_t,s-t)
\end{equation}     
  
We can cleverly use Euler's formula \(e^{ix}=cosx+isinx\), where the real part is cosx and the imaginary part sinx is of a complex number.

\begin{table*}[h]
\centering
\caption{Comparison of experimental results of various methods under different missing data conditions. The best performances are presented in bold and the second best performances are presented in underline.}
\adjustbox{width=\linewidth}{
\begin{tabular}{c|ccccc|ccccc|ccccc}
\toprule
\multirow{2}{*}{Methods} & \multicolumn{5}{c|}{16\% missing normal, 4\% missing EBV} & \multicolumn{5}{c|}{20\% missing EBV} & \multicolumn{5}{c}{20\% missing normal} \\
\cmidrule(lr){2-6} \cmidrule(lr){7-11} \cmidrule(lr){12-16}
& Accuracy & F1-Score & Recall & AUC & Precision & Accuracy & F1-Score & Recall & AUC & Precision & Accuracy & F1-Score & Recall & AUC & Precision \\
\midrule
GNN4CMR \cite{Qian2022IntegratingMC} & 0.73 & 0.32 & 0.16 & 0.59 & 0.58 & 0.75 & 0.31 & 0.25 & 0.63 & 0.53 & 0.68 & 0.26 & 0.12 & 0.65 & 0.62 \\
MMCL \cite{Hager2023BestOB} & 0.71 & 0.28 & \textbf{0.24} & 0.64 & 0.60 & 0.78 & 0.27 & 0.27 & 0.66 & 0.61 & 0.71 & 0.18 & 0.15 & 0.59 & 0.56 \\
HGCN  \cite{Hou2023HybridGC} & 0.69 & 0.32 & 0.19 & 0.63 & 0.48 & 0.76 & 0.24 & 0.19 & 0.62 & 0.41 & 0.76 & \textbf{0.29} & 0.14 & 0.68 & 0.63 \\
RBA-GCN \cite{Yuan2023RBAGCNRB} & 0.64 & 0.26 & 0.2 & 0.62 & 0.32 & 0.81 & 0.28 & 0.21 & 0.58 & 0.25 & 0.74 & 0.25 & \textbf{0.18} & 0.61 & 0.71 \\
MPMM  \cite{bib13}& 0.80 & 0.27 & 0.21 & 0.71 & 0.32 & 0.83 & 0.25 & \textbf{0.38} & 0.72 & 0.18 & 0.91 & 0.21 & 0.13 & 0.71 & 0.64 \\
\textbf{Ours} & \textbf{0.94} & \textbf{0.35} & \underline{0.23} & \textbf{0.89} & \textbf{0.75} & \textbf{0.94} & \textbf{0.35} & \underline{0.31} & \textbf{0.92} & \textbf{0.75} & \textbf{0.94} & \underline{0.27} & \underline{0.16} & \textbf{0.84} & \textbf{0.88} \\
\bottomrule
\end{tabular}
}
\label{tab:combined}
\end{table*}

After transformation, formulas \textbf{O} and  \textbf{g} can be changed to:
\begin{align}     
  &e^{is\theta}=cos(s\theta)+isin(s\theta)\\      
  &e^{it\theta}=cos(t\theta)+isin(t\theta)      \\
  &e^{i(s-t)\theta}=cos((s-t)\theta)+isin((s-t)\theta)     \\ 
  &O_q(x_s,s)=(W_qx_s)e^{is\theta}
\end{align} 

Then, according to linear algebra, we can represent \(q_s\) using a matrix:
\begin{align}       
  &q_s=\begin{pmatrix} q_s^{(1)} \\ q_s^{(2)} \end{pmatrix}=(W_qx_s)=\begin{pmatrix} W_q^{(11)} & W_q^{(12)} \\ W_q^{(21)} & W_q^{(22)}\end{pmatrix}\begin{pmatrix} x_s^{(1)} \\ x_s^{(2)} \end{pmatrix}\\   
  &O_q(x_s,s)=(W_qx_s)e^{is\theta}=q_se^{is\theta}
\end{align}

Therefore, multiplying these two complex numbers, we get the following result:
\begin{align}
q_se^{is\theta}=[q_s^{(1)}cos(s\theta)-q_s^{(2)}sin(s\theta) ,q_s^{(2)}cos(s\theta)+q_s^{(1)}sin(s\theta)
\end{align} 
     
Subsequently, we find that the above expression is equivalent to multiplying the query vector by a rotation matrix:
\begin{align}
  P_q(x_s,s) & = (W_qx_s)e^{is\theta} = q_se^{is\theta} \nonumber \\
            & = \begin{pmatrix} \cos(s\theta) & -\sin(s\theta) \\ \sin(s\theta) & \cos(s\theta)\end{pmatrix}\begin{pmatrix} q_s^{(1)} \\ q_s^{(2)}\end{pmatrix}
\end{align}

Similarly, the key vector \(k_t\) can be represented as follows:
\begin{align}
  P_k(x_t,t) &= (W_kx_t)e^{in\theta} = k_te^{it\theta} \nonumber \\
             & = \begin{pmatrix} \cos(t\theta) & -\sin(t\theta) \end{pmatrix} \begin{pmatrix} k_t^{(1)} \\ k_t^{(2)} \end{pmatrix} \nonumber \\
             &\quad + \begin{pmatrix} \sin(t\theta) & \cos(t\theta)\end{pmatrix}\begin{pmatrix} k_t^{(2)} \\ k_t^{(1)} \end{pmatrix}
\end{align}

By rearranging the above formulas, we can simplify the following expression:
\begin{align}
&<P_q(x_s,s),P_k(x_t,t)> \nonumber \\
&= \left(\begin{pmatrix} \cos(s\theta) & -\sin(s\theta) \\ \sin(s\theta) & \cos(s\theta)\end{pmatrix}^T \begin{pmatrix} q_s^{(1)} \\ q_s^{(2)} \end{pmatrix}\right)^T \nonumber \\
&\quad \begin{pmatrix} \cos(t\theta) & -\sin(t\theta) \\ \sin(t\theta) & \cos(t\theta)\end{pmatrix}\begin{pmatrix} k_t^{(1)} \\ k_t^{(2)} \end{pmatrix} \nonumber \\
&= \begin{pmatrix} q_s^{(1)} & q_s^{(2)} \end{pmatrix}\begin{pmatrix} \cos((s-t)\theta) & -\sin((s-t)\theta) \\ \sin((s-t)\theta) & \cos((s-t)\theta)\end{pmatrix}\begin{pmatrix} k_t^{(1)} \\ k_t^{(2)} \end{pmatrix}
\end{align}
  
In essence, the self-attention process with SCAI involves several steps: first, calculating query and key vectors for each feature embedding in the token sequence; next, computing rotated position embeddings for each token position; then, applying rotation transformations to the query and key vectors pair-wise; and finally, calculating the dot product between them to derive the self-attention result. The SCAI module addresses the performance challenges of traditional Transformers with high-dimensional, multi-modal medical data, particularly in scenarios with missing key modalities. By integrating SCAI, IMAN optimizes feature extraction adaptively, ensuring robust prediction accuracy even when certain data types are missing.

\subsection{Context-Aware Feature Acquisition Module}

Existing methodologies for extracting multi-modal image characteristics, as cited in \cite{bib13,bib24}, are limited by conventional convolution's localized receptive field, which restricts effective capture of global contextual information. Drawing inspiration from Zhang's pioneering work \cite{Zhang2023AKConvCK}, we introduce the Context-Aware Feature Acquisition (CAFA) module. This module innovatively overcomes these constraints by dynamically adjusting convolution kernel positions through trainable offsets. It enables adaptive feature extraction across diverse scales and orientations in multi-modal medical imagery.

Given the input image $I \in \mathbb{R}^{C \times H \times W}$, where $C$, $H$, and $W$ denote the number of channels, height, and width respectively. The CAFA generates an initial sampling coordinate $P_n$ via Algorithm \ref{algorithm:AFE}. Then $I$ undergoes a convolution operation to learn the offset, $\text{offset} = P_{\text{conv}}(I)$. The offset is used to adjust the shape of the sample at each position. The adjusted sampling position $P$ is mathematically defined as:
\begin{equation}
    P=P_0+P_n+\text{offset}
\end{equation}
where $P_0$ denotes the base grid coordinates.

Next, the CAFA module will resample the input features at position $P$ using bilinear interpolation. The resampled input feature can be expressed as:
\begin{equation}
    x_{\mathrm{offset}}=g_{lt}\cdot x_{q_{lt}}+g_{rb}\cdot x_{q_{rb}}+g_{lb}\cdot x_{q_{lb}}+g_{rt}\cdot x_{q_{rt}}
\end{equation}
where $g_{lt}$, $g_{rb}$, $g_{lb}$, $g_{rt}$ are the weights of the bilinear interpolation, and $x_{q_{lt}}$, $x_{q_{rb}}$, $x_{q_{lb}}$, $x_{q_{rt}}$ are the eigenvalues of the corresponding positions. lt, lb, rt, rb are abbreviations for left-top, left-bottom, right-top, and right-bottom, respectively.

Finally, the CAFA reshapes the resampled features into the desired shape and obtains the output feature map by a depthwise  convolution operation:
\begin{equation}
    output=Dconv(x_{\mathrm{offset}})
\end{equation}
\begin{figure*}[ht]
  \centering
  \footnotesize
  \begin{tabular}{ccc}
    \includegraphics[width=0.32\linewidth]{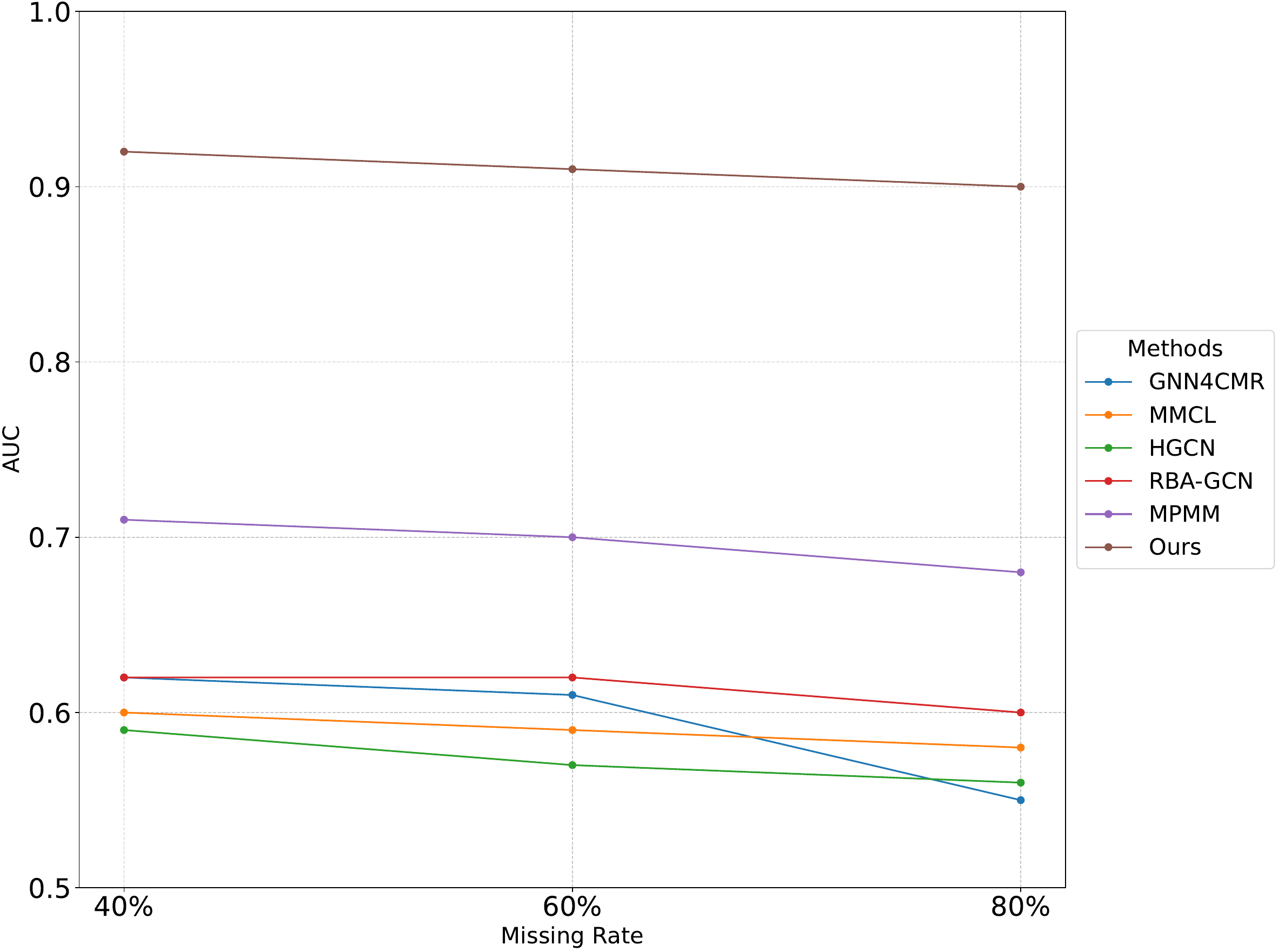}&
    \includegraphics[width=0.32\linewidth]{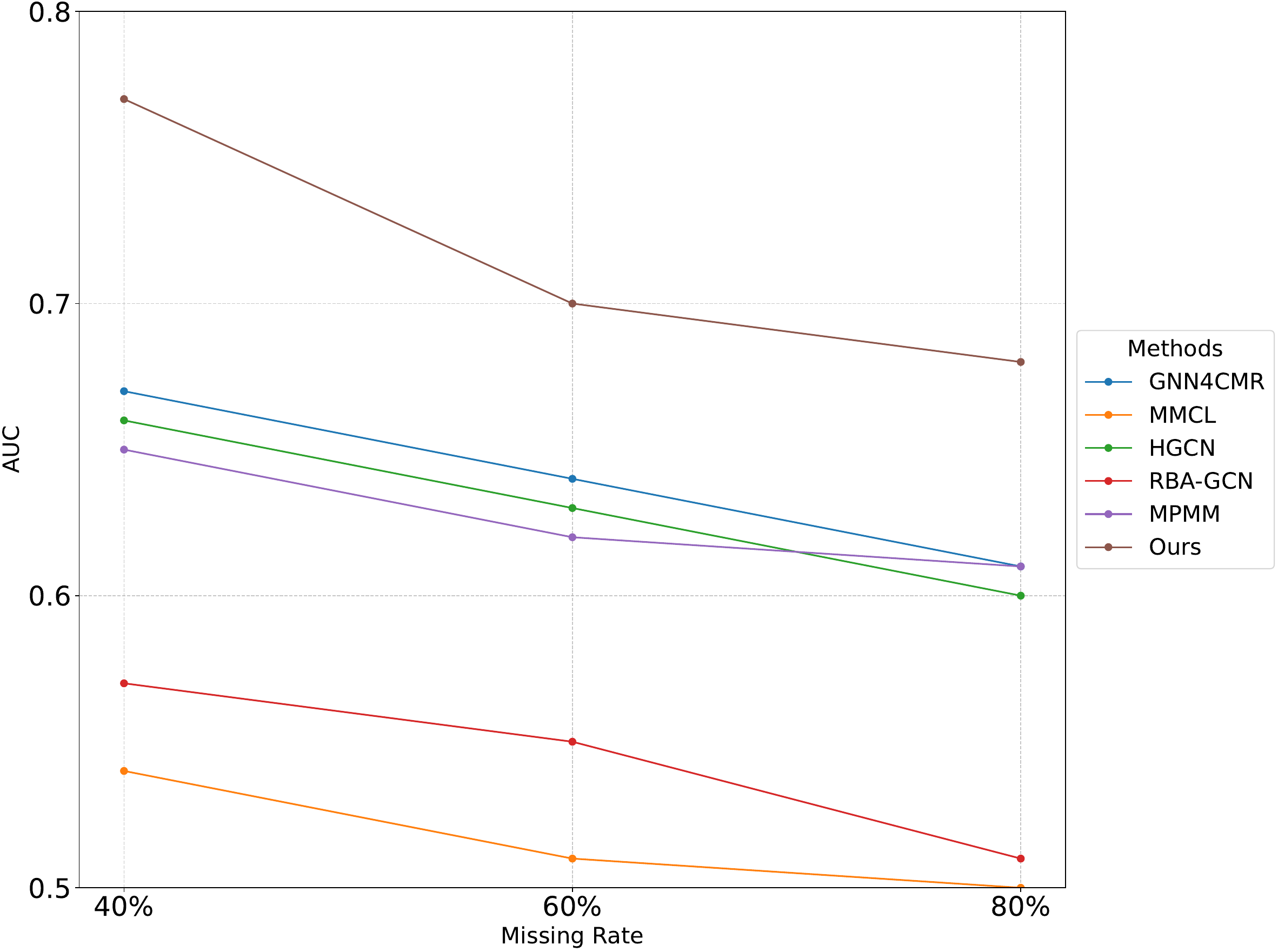} &  
  \includegraphics[width=0.32\linewidth]{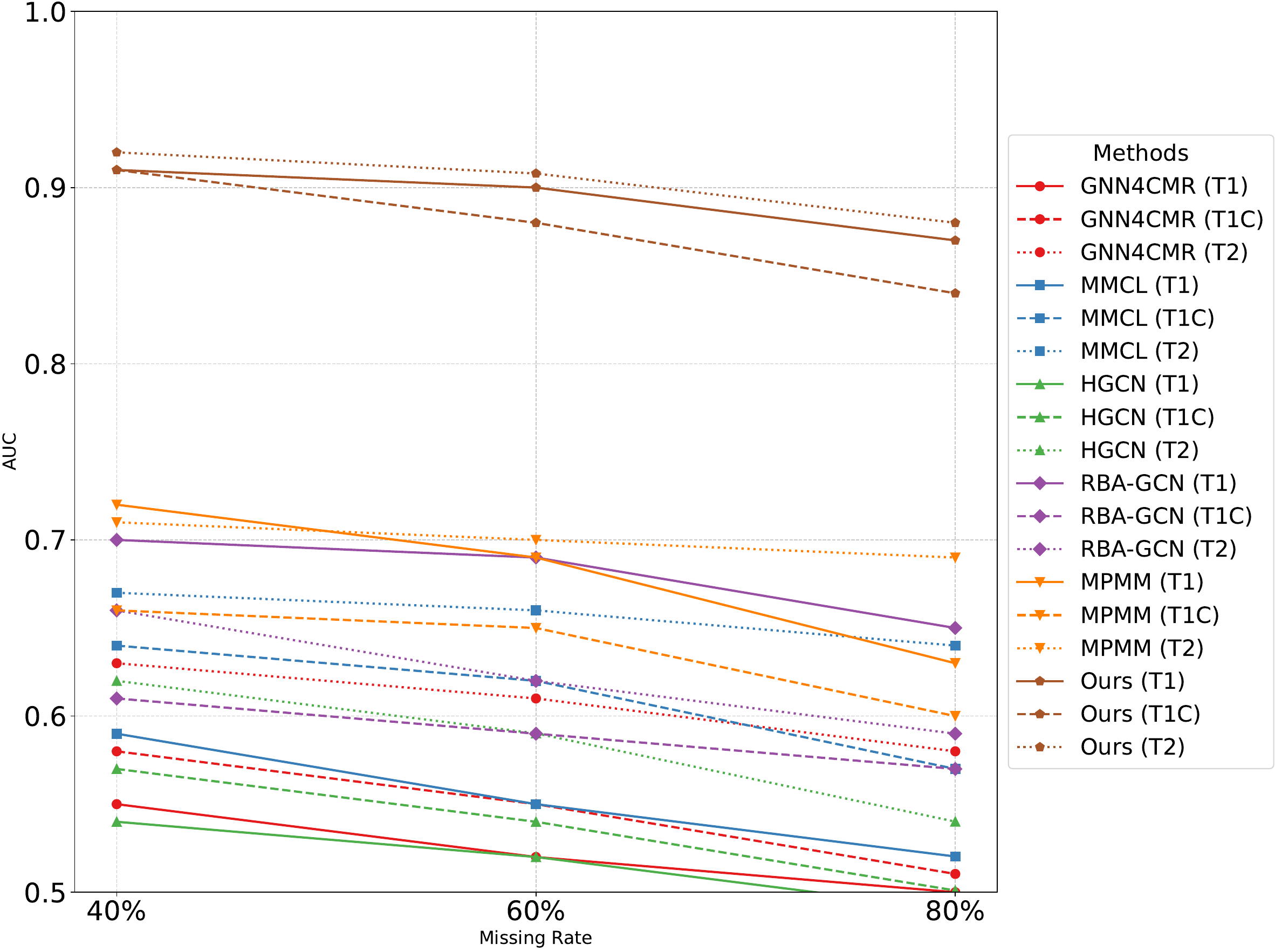}
     \\
    
  \end{tabular}
  \caption{Comparison of AUC performance of various methods with different missing rates: EBV, ``normal'', and medical imaging modalities.}
  \label{fig:loss}
\end{figure*}
Through this mechanism, the CAFA module endows convolution kernels with the capacity to incorporate a variable number of parameters, dynamically tuning the sampling grid to conform to the evolving characteristics of the target. By tailoring optimal sampling configurations to the specific task at hand, the CAFA module significantly augments its capacity to discern salient features across a broad spectrum of scales within multi-modal medical images, thereby enhancing the model's discriminative power and generalizability.



\section{Experiment}

\subsection{Datasets and Evaluation Metric}
The dataset for this study was provided by Sun Yat-sen University Cancer Center and associated research institutions (approved ID: B2019-222-01). 
It comprises 1,224 samples divided into training (70\%), validation (15\%), and test (15\%) sets.  Five modalities were considered: two diagnostic fields (EBV and normal) and three medical imaging modalities (T1, T1C, and T2). The EBV field is represented by a 4-dimensional vector, and the normal field by a 19-dimensional vector. Each imaging modality is a $32\times 224\times 224$ tensor. To simulate real-world data incompleteness, we implemented a sophisticated missingness mechanism using a missingness table $\mathbf{T}$, a $1224\times5$ binary matrix where each entry indicates the presence (0) or absence (1) of data. The missingness rate for a modality \(k\) is computed as $\sum(\mathbf{T}[:, k]) / \mathbf{T} . \text {shape}[0]$ (where \(k=0, 1, 2, 3, 4\)). Our experiments will explore overall missing rates of 20\%, 40\%, 60\% and 80\%, focusing primarily on the 20\% rate, which reflects real medical scenarios according to specialized doctors. 
We evaluated our approach using Accuracy, F1-Score, Recall, AUC, and Precision.
\subsection{Experiment details}

\begin{table}[ht]
\centering
\caption{Comparison of methods under different missing data conditions. 
}
\adjustbox{width=\columnwidth}{
\begin{tabular}{l|cc|cc|cc|cc}
\toprule
\multirow{2}{*}{Method} & \multicolumn{2}{c|}{16\% normal, 4\% T1} & \multicolumn{2}{c|}{20\% T1} & \multicolumn{2}{c|}{20\% T1C} & \multicolumn{2}{c}{20\% T2} \\
\cmidrule(lr){2-3} \cmidrule(lr){4-5} \cmidrule(lr){6-7} \cmidrule(lr){8-9}
& Accuracy & AUC & Accuracy & AUC & Accuracy & AUC & Accuracy & AUC \\
\midrule
GNN4CMR \cite{Qian2022IntegratingMC} & 0.67 & 0.59 & 0.75 & 0.52 & 0.71 & 0.58 & 0.73 & 0.62 \\
MMCL \cite{Hager2023BestOB} & 0.71 & 0.66 & 0.79 & 0.62 & 0.72 & 0.66 & 0.68 & 0.69 \\
HGCN \cite{Hou2023HybridGC} & 0.74 & 0.64 & 0.71 & 0.54 & 0.65 & 0.59 & 0.70 & 0.62 \\
RBA-GCN \cite{Yuan2023RBAGCNRB} & 0.70 & 0.68 & 0.73 & 0.61 & 0.70 & 0.63 & 0.76 & 0.68 \\
MPMM \cite{bib13} & 0.82 & 0.71 & 0.81 & 0.72 & 0.78 & 0.69 & 0.75 & 0.73 \\
\textbf{Ours} & \textbf{0.93} & \textbf{0.84} & \textbf{0.93} & \textbf{0.92} & \textbf{0.93} & \textbf{0.92} & \textbf{0.94} & \textbf{0.90} \\
\bottomrule
\end{tabular}
}
\label{tab:combined}
\end{table}

Our approach was trained on a single RTX 4090 GPU and the pytorch framework. Where the batch is set to 64, the base backbone is ViLT \cite{bib24}, the Adam optimizer is used, the initial learning rate is set to 1e-4, the weight decay coefficient is set to 0.01, and the epoch is set to 200. The learning rate is warmed up for $10\%$ of the total training steps and then is decayed linearly to zero.

\subsection{Comparison with State-of-the-Art Methods}

\begin{table}[ht]
  \centering
  \caption{Ablation Study on each module.}
    \begin{tabular}{ccc|cc}
    \toprule
    \multicolumn{3}{c|}{Module} & \multicolumn{2}{c}{\textbf{Metrics}} \\
    \midrule
    DCMC & SCAI & CAFA   & \multicolumn{1}{c}{\textbf{Accuracy $\uparrow$}} & \multicolumn{1}{c}{\textbf{AUC $\uparrow$}} \\
    \midrule
    \xmark & \xmark & \xmark & \multicolumn{1}{c}{0.83} & \multicolumn{1}{c}{0.72} \\
    \checkmark & \xmark & \xmark & \multicolumn{1}{c}{0.92} & \multicolumn{1}{c}{0.80} \\
    \checkmark & \checkmark & \xmark & 0.93 & 0.91 \\
    
     \checkmark & \checkmark & \checkmark & \multicolumn{1}{c}{\textbf{0.94}} & \multicolumn{1}{c}{\textbf{0.92}} \\
    \bottomrule
    \end{tabular}%
    
  \label{ab}
\end{table}%




Our comprehensive experiments comparing IMAN with state-of-the-art methods like GNN4CMR \cite{Qian2022IntegratingMC}, MMCL \cite{Hager2023BestOB}, HGCN \cite{Hou2023HybridGC}, RBA-GCN \cite{Yuan2023RBAGCNRB}, and MPMM \cite{bib13} across various missing data scenarios demonstrate IMAN's superior performance, particularly in Accuracy and AUC (see Table \ref{tab:combined}). For example, with 20\% missing EBV data, IMAN achieved an accuracy of 0.94 and an AUC of 0.92, significantly outperforming MPMM's accuracy of 0.83 and AUC of 0.72. However, the relatively lower F1-Score and Recall metrics may be due to the inherent class imbalance in nasopharyngeal carcinoma datasets, where negative cases typically outnumber positive ones. Additionally, IMAN may reflect a conservative diagnostic approach, leading to fewer false positives but potentially more false negatives. The complexity of cancer progression, influenced by factors like genetic expression and treatment response, may result in some rare predictive features being overlooked. Furthermore, predicting long-term mortality poses challenges in capturing all relevant factors. While IMAN excels in handling missing data, the absence of certain key predictive factors may still impact its performance in identifying all positive cases. The model's design, focusing on optimizing accuracy and AUC, might have led to trade-offs in F1-Score and Recall. Despite these limitations, IMAN's overall performance surpasses existing methods, especially in scenarios with incomplete data, highlighting its potential to enhance the accuracy and reliability of mortality predictions in clinical settings for nasopharyngeal carcinoma as show in Fig. \ref{fig:loss}.

\subsection{Ablation Study}

To thoroughly evaluate the contribution and effectiveness of each module in the IMAN network, we conducted a series of detailed ablation experiments. These experiments were designed to assess the impact of the DCMC, SCAI, and CAFA modules on the model's overall performance. The experimental setup simulated common clinical scenarios with $20\%$ of data missing EBV-related information. Table \ref{ab} presents the results of the experiments with various module combinations.

\section{Conclusion}
In this study, we introduce IMAN, an adaptive network for predicting mortality in nasopharyngeal carcinoma cases with missing modalities. IMAN includes several key modules: the DCMC module normalizes heterogeneous inputs by scaling medical images and field data using learnable parameters, the SCAI module enhances multi-modal feature fusion with positional information in a self-attention mechanism, and the CAFA module adjusts convolution kernel positions for adaptive feature capture. These components work together to improve the precision of the system in handling medical data, enabling the detection of subtle patterns and effective feature capture across various scales and orientations. Experimental results in different missing data scenarios show that our method outperforms current alternatives. Future work will explore the performance of our model under varying data missing rates and more complex missing data scenarios.

\section*{Acknowledgment}
This work was partially supported by the Guangzhou Key Areas Research and Development Program (Grant 2023B01J0029) and the Guangdong Provincial Key Laboratory of Cyber-Physical Systems (Grant 2020B1212060069), in part by the University of Macau (Grant MYRG2022-00190-FST) and the Science and Technology Development Fund of Macau SAR (Grants 0141/2023/RIA2 and 0193/2023/RIA3).

\section*{Ethical Statement}
This study was conducted in accordance with ethical guidelines and was approved by the Ethics Committee of Sun Yat-sen University Cancer Center IRB, approval number: B2019-222-01, approval date: January 8, 2020. We ensured that the privacy and confidentiality of the participants was strictly maintained. All data collected were anonymized and stored securely. Only authorized personnel had access to the data and measures were taken to prevent any unauthorized use or disclosure. The authors declare that they have no conflict of interest.

\bibliographystyle{ieeetr} 
\bibliography{reference}

\begin{thebibliography}{10}

\bibitem{bib4}
J.~Sheng, S.~Lam, T.~Zhou, J.~Zhang, Y.~Zhang, and J.~Cai, ``Multi-kernel fusion with fuzzy label relaxation for predicting distant metastasis in nasopharyngeal carcinoma,'' in {\em DSP}, pp.~1--5, 2023.

\bibitem{py1}
J.~Zhang, H.~Li, D.~Xu, Y.~Lou, M.~Ran, Z.~Jin, and Y.~Huang, ``Decouple and decorrelate: A disentanglement security framework combing sample weighting for cross-institution biased disease diagnosis,'' {\em IEEE Internet of Things Journal}, 2024.

\bibitem{py2}
J.~Zhang, D.~Xu, Y.~Lou, and Y.~Huang, ``A novel multi-atlas fusion model based on contrastive learning for functional connectivity graph diagnosis,'' in {\em ICASSP}, pp.~2295--2299, 2024.

\bibitem{py4}
Y.~Lou, J.~Zhang, D.~Xu, Y.~Cao, H.~Wang, and Y.~Huang, ``No-reference mri quality assessment via contrastive representation: Spatial and frequency domain perspectives,'' in {\em ICME}, pp.~1--6, 2024.

\bibitem{py5}
Y.~Lou, Y.~Chen, D.~Xu, D.~Zhou, Y.~Cao, H.~Wang, and Y.~Huang, ``Refining the unseen: Self-supervised two-stream feature extraction for image quality assessment,'' in {\em ICDM}, pp.~1193--1198, 2023.

\bibitem{py6}
R.~Zhang, Y.~Lou, D.~Xu, Y.~Cao, H.~Wang, and Y.~Huang, ``A learnable discrete-prior fusion autoencoder with contrastive learning for tabular data synthesis,'' in {\em AAAI}, vol.~38, pp.~16803--16811, 2024.

\bibitem{bib6}
R.~Tibshirani, ``Regression shrinkage and selection via the lasso,'' {\em Journal of the royal statistical society series b-methodological}, vol.~58, pp.~267--288, 1996.

\bibitem{chen1}
X.~Chen, C.-M. Pun, and S.~Wang, ``Medprompt: Cross-modal prompting for multi-task medical image translation,'' {\em arXiv}, 2023.

\bibitem{py3}
D.~Xu, Y.~Chen, J.~Zhang, Y.~Lou, H.~Wang, J.~He, and Y.~Huang, ``Radiology report generation via structured knowledge-enhanced multi-modal attention and contrastive learning,'' in {\em BIBM}, pp.~2320--2325, 2023.

\bibitem{bib7}
Y.~Zhang, N.~He, J.~Yang, Y.~Li, D.~Wei, Y.~Huang, Y.~Zhang, Z.~He, and Y.~Zheng, ``mmformer: Multimodal medical transformer for incomplete multimodal learning of brain tumor segmentation,'' {\em ArXiv}, 2022.

\bibitem{bib8}
J.~Chen and A.~Zhang, ``Hgmf: Heterogeneous graph-based fusion for multimodal data with incompleteness,'' in {\em ACM SIGKDD}, 2020.

\bibitem{bib9}
T.~Wang, W.~Zhu, Y.~Gao, J.~Feng, and S.~Zhang, ``Hgcn: Harmonic gated compensation network for speech enhancement,'' in {\em ICASSP}, pp.~371--375, 2022.

\bibitem{bib10}
S.~Parthasarathy and S.~Sundaram, ``Training strategies to handle missing modalities for audio-visual expression recognition,'' in {\em ICMI}, 2020.

\bibitem{chen3}
L.~Zhu, W.~Liu, X.~Chen, Z.~Li, X.~Chen, Z.~Wang, and C.-M. Pun, ``Test-time intensity consistency adaptation for shadow detection,'' {\em arXiv}, 2024.

\bibitem{chen5}
X.~Li, G.~Huang, L.~Cheng, G.~Zhong, W.~Liu, X.~Chen, and M.~Cai, ``Cross-domain visual prompting with spatial proximity knowledge distillation for histological image classification,'' {\em Journal of Biomedical Informatics}, p.~104728, 2024.

\bibitem{bib24}
W.~Kim, B.~Son, and I.~Kim, ``Vilt: Vision-and-language transformer without convolution or region supervision,'' in {\em International Conference on Machine Learning}, 2021.

\bibitem{Tikhonova2023TextSA}
M.~Tikhonova and A.~Fenogenova, ``Text simplification as a controlled text style transfer task,'' {\em Computational Linguistics and Intellectual Technologies}, 2023.

\bibitem{Su_Lu_Pan_Wen_Liu_2021}
J.~Su, Y.~Lu, S.~Pan, B.~Wen, and Y.~Liu, ``Roformer: Enhanced transformer with rotary position embedding.,'' {\em arXiv}, 2021.

\bibitem{chen2}
F.~Zheng, X.~Chen, W.~Liu, H.~Li, Y.~Lei, J.~He, C.-M. Pun, and S.~Zhou, ``Smaformer: Synergistic multi-attention transformer for medical image segmentation,'' in {\em BIBM}, 2024.

\bibitem{chen4}
X.~Guo, X.~Chen, S.~Luo, S.~Wang, and C.-M. Pun, ``Dual-hybrid attention network for specular highlight removal,'' in {\em ACM MM}, 2024.

\bibitem{Qian2022IntegratingMC}
S.~Qian, D.~Xue, Q.~Fang, and C.~Xu, ``Integrating multi-label contrastive learning with dual adversarial graph neural networks for cross-modal retrieval,'' {\em IEEE Transactions on Pattern Analysis and Machine Intelligence}, vol.~45, pp.~4794--4811, 2022.

\bibitem{Hager2023BestOB}
P.~Hager, M.~J. Menten, and D.~Rueckert, ``Best of both worlds: Multimodal contrastive learning with tabular and imaging data,'' {\em CVPR}, pp.~23924--23935, 2023.

\bibitem{Hou2023HybridGC}
W.~Hou, C.~Lin, L.~Yu, J.~Qin, R.~Yu, and L.~Wang, ``Hybrid graph convolutional network with online masked autoencoder for robust multimodal cancer survival prediction,'' {\em IEEE Transactions on Medical Imaging}, vol.~42, pp.~2462--2473, 2023.

\bibitem{Yuan2023RBAGCNRB}
L.~Yuan, G.~Huang, F.~Li, X.~Yuan, C.-M. Pun, and G.~Zhong, ``Rba-gcn: Relational bilevel aggregation graph convolutional network for emotion recognition,'' {\em IEEE/ACM Transactions on Audio, Speech, and Language Processing}, vol.~31, pp.~2325--2337, 2023.

\bibitem{bib13}
Y.-L. Lee, Y.-H. Tsai, W.-C. Chiu, and C.-Y. Lee, ``Multimodal prompting with missing modalities for visual recognition,'' {\em CVPR}, pp.~14943--14952, 2023.

\bibitem{Zhang2023AKConvCK}
X.~Zhang, Y.~Song, T.~Song, D.~Yang, Y.~Ye, J.~Zhou, and L.~Zhang, ``Akconv: Convolutional kernel with arbitrary sampled shapes and arbitrary number of parameters,'' {\em ArXiv}, 2023.

\end{thebibliography}

\end{document}